\documentclass{article}

\usepackage{arxiv}

\usepackage[utf8]{inputenc} % allow utf-8 input
\usepackage[T1]{fontenc}    % use 8-bit T1 fonts
\usepackage{hyperref}       % hyperlinks
\usepackage{url}            % simple URL typesetting
\usepackage{booktabs}       % professional-quality tables
\usepackage{amsfonts}       % blackboard math symbols
\usepackage{nicefrac}       % compact symbols for 1/2, etc.
\usepackage{microtype}      % microtypography
\usepackage{lipsum}		% Can be removed after putting your text content
\usepackage{graphicx}
\usepackage{natbib}
\usepackage{doi}

\usepackage{latexsym,amsmath,amssymb}

\usepackage{lipsum}% http://ctan.org/pkg/lipsum
\usepackage{multicol}% http://ctan.org/pkg/multicols
\usepackage{graphicx}% http://ctan.org/pkg/graphicx

\newtheorem{proposition}{Proposition}
\newtheorem{definition}{Definition}
\newtheorem{lemma}{Lemma}

\def\bfx{{\bf x}} 
\def\bfr{{\bf r}} \def\bfs{{\bf s}}
 
 \def\bfa{{\bf a}}

\title{Sequential Decision Making on Unmatched Data using Bayesian Kernel Embeddings}

\date{} 					% Or removing it

\author{Diego ~Martinez-Taboada\thanks{Work primarily done at the University of Oxford and finished at Carnegie Mellon University.} \\
	Department of Statistics \& Data Science\\
	Carnegie Mellon University\\
	\texttt{diegomar@andrew.cmu.edu} \\
	\And
	Dino ~Sejdinovic \\
	Department of Statistics\\
	University of Oxford \\
	\texttt{dino.sejdinovic@stats.ox.ac.uk} \\
}

\renewcommand{\undertitle}{Preprint. Under review.}
\renewcommand{\shorttitle}{\textit{arXiv} Template}

\hypersetup{
pdftitle={Sequential Decision Making on Unmatched Data using Bayesian Kernel Embeddings},
pdfauthor={Diego ~Martinez-Taboada, Dino ~Sejdinovic},
}

\begin{document}
\maketitle

\begin{abstract}
	  The problem of sequentially maximizing the expectation of a function seeks to maximize the expected value of a function of interest without having direct control on its features. Instead, the distribution of such features depends on a given context and an action taken by an agent. In contrast to Bayesian optimization, the arguments of the function are not under agent's control, but are indirectly determined by the agent's action based on a given context. If the information of the features is to be included in the maximization problem, the full conditional distribution of such features, rather than its expectation only, needs to be accounted for. Furthermore, the function is itself unknown, only counting with noisy observations of such function, and potentially requiring the use of unmatched data sets. We propose a novel algorithm for the aforementioned problem which takes into consideration the uncertainty derived from the estimation of both the conditional distribution of the features and the unknown function, by modeling the former as a Bayesian conditional mean embedding and the latter as a Gaussian process. Our algorithm empirically outperforms the current state-of-the-art algorithm in the experiments conducted.
\end{abstract}

\section{INTRODUCTION}

Uncertainty quantification is inherent to sequential decision making problems, where an agent sequentially explores a set of possible actions while seeking to maximize the associated reward (\cite{alagoz2010markov}, \cite{sutton2018reinforcement}). The algorithms designed for such problems are based on an exploration-exploitation trade-off: the agent tries to exploit its knowledge on the actions that have yielded high rewards, but it also seeks to explore actions that count with little information (\cite{macready1998bandit}, \cite{audibert2009exploration}). 

The multi-armed bandit (MAB) is one of the sequential decision problems that has received the most attention in the literature (\cite{vermorel2005multi}, \cite{slivkins2019introduction}). The classical MAB problem is defined by a tuple $\langle \mathcal{A}, F_R \rangle$, where $\mathcal{A}$ is the state of actions and $F_R$ the reward distribution. At time step $t \in \mathbb{N}$, the agent decides to take an action $a_t \in \mathcal{A}$. A reward $r_{a_t}$ drawn from $r_{a_t} \sim F_R(\cdot | a_t)$ follows. The MAB aims at maximizing the cumulative reward
\begin{equation}
    J_T = \mathbb{E}\left[\sum_{t = 1}^T r_{a_t}\right],
\end{equation}
where $a_t$ are the actions sequentially taken. The contextual multi-armed bandit problem (CMAB) generalizes the MAB by including a state space (or context space) $\mathcal{S}$ (\cite{lu2010contextual}, \cite{zhou2015survey}). Formally, the CMAB problem is determined by a tuple $\langle \mathcal{S}, \mathcal{A}, F_R \rangle$, and it aims at maximizing 
\begin{equation}
    J_T = \mathbb{E}\left[ \sum_{t = 1}^T r_{s_t, a_t} \right],
\end{equation}
where $r_{s_t, a_t} \sim F_R(\cdot | s_t, a_t)$, and $s_t$ are sequentially given.

However, one may be interested in optimizing the expectation of a function $f(r)$, instead of $r$ itself. In other words, $r_t$ may be seen as an intermediate reward that mediates between the agent's actions $a_t$ and the ultimate reward $f(r_t)$. In various fields of application, it is usual to study a function of the outcome, rather than the outcome itself. For instance, the conditional value at risk (CVaR), which has seen extensive use in financial portfolio optimization (\cite{rockafellar2000optimization}, \cite{zhu}), considers $f(r) = r 1_{r \leq r_\alpha}$. The median and other quantiles may be of interest for noisy data (\cite{even2002pac}, \cite{altschuler}). 

In more general settings, the function $f$ may be itself unknown. Noisy data of the form
\begin{equation}
    (r_i, f(r_i) + \xi_i)_{i = 1}^M
\end{equation}
may be the only available information on such dependency. 
In this case, uncertainty raises from the lack of information between the traditional CMAB variables $r, a, s$, but also from the unknown function $f$. 

Please note that this scenario opens the door for considering multiple sources of data. In contrast to the usual \textit{matched} scenario, where the sequential data is of the form $(s_i, a_i, r_i, f(r_i) + \xi_i)_{i = 1}^t$, it may be of interest to consider an \textit{unmatched} scenario, where a second data set of the form $(r_j, f(r_j) + \xi_j)_{j = 1}^M$ is given on top of the sequential data. Algorithms that handle unmatched data sets have currently been attracting more and more interest in the so called data fusion problem (\cite{meng2020survey}).

For instance, a new movie recommendation system might obtain a multivariate $r_j$ consisting on: whether the user showed interest in the recommendation, the number of cliques after the recommendation screen, the time spent reading the description of the recommendation, etc. The final reward $f(r_j) + \xi_j$ is given by the number of minutes that the client ended up watching the content for. A rich data set $(r_j, f(r_j) + \xi_j)_{j = 1}^M$ might be obtained from other platforms that have been longer in the market.
Not considering this second data set would imply a substantial loss of information which would to suboptimal approaches.

In this work, we revisit the problem of sequentially maximising the conditional expectation of a function in the CMAB framework. Our contributions are two-folded:
\begin{itemize}
    \item We design a novel algorithm, namely Contextual Bayesian Mean Process Upper Confidence Bound, which considers two sources of uncertainty derived from the lack of information on $r, a, s$, as well as the unknown function $f$. It allows for considering both matched and unmatched data sets.
    \item We empirically show that the Bayesian Mean Process Upper Confidence Bound outperforms the state-of-the-art Conditional Mean Embeddings Upper Confidence Bound (\cite{chowdhury2020active}) in the experimental settings considered.
\end{itemize}

\section{RELATED WORK}

Multiple algorithms have been proposed for addressing the CMAB problem. The \textit{contextual Gaussian process upper confidence bound} (CGP-UCB), which motivates the algorithm proposed in this work, was first presented in \cite{krause2011contextual}. It generalizes the context-free \textit{Gaussian process upper confidence bound} (GP-UCB) introduced in \cite{srinivas2009gaussian}. The CGP-UCB attains sublinear contextual regret in many real life applications i.e. it is able to compete with the optimal mapping from contexts to actions. However, please note that there exist prominent alternatives to CGP-UCB such as Thompson sampling (\cite{agrawal2013thompson}). Some algorithms designed for the MAB problem (\cite{chowdhury2017kernelized}) may also be adapted to the CMAB framework.

The problem of sequentially maximising the conditional expectation has not received as much attention in the literature, although a number of closely related problems have been presented in different forms. \cite{oliveira2019bayesian} proposed a sequential decision problem where both the reward and actions are noisy (execution and localisation noise), with applications in robotics and stochastic simulations. The contextual combinatorial multi-armed bandit problem (\cite{chen2013combinatorial}, \cite{qin2014contextual}) may be understood as the problem of sequentially maximixing the conditional expectation of a function, where the chosen super arm may be seen as a multivariate binary action, and the dependency between the scores of the individual arms and the reward as the unknown function.

For the problem of sequentially maximizing the expectation of a function, \cite{chowdhury2020active} proposed the Conditional Mean Embeddings Upper Confidence Bound (CME-UCB) algorithm. The CME-UCB uses the conditional mean embedding for modeling the conditional distribution of the features. In contrast, alternative approaches for such conditional density estimation scale poorly with the dimension of the underlying space (\cite{grunewalder2012modelling}). 

The conditional mean embedding, key element of the CME-UCB algorithm, extends the concept of kernel mean embeddings to conditional distributions (\cite{muandet2017kernel}). In terms of Bayesian procedures, \cite{flaxman_bayesian_kernel_embedding} proposed a Bayesian approach for learning mean embeddings, and \cite{bayesimp} generalized the approach for Bayesian conditional mean embeddings (BayesCME). Such Bayesian learning of mean embeddings requires the concept of nuclear dominance (\cite{sample_path_rkhs}), which allows for defining a GP with trajectories in a Reproducing Kernel Hilbert Space with probability 1. 

Furthermore, conditional mean processes are now well established (\cite{chau2021deconditional}), which study the integral of a Gaussian process (GP) with respect to a conditional distribution. Building on the Bayesian conditional mean embedding and the conditional mean process, \cite{bayesimp} proposed the BayesIMP algorithm for tackling a data fusion problem in a causal inference setting. Such algorithm considers uncertainty derived from two data samples, with observations of a mediating variable included in both data samples.

\section{BACKGROUND}

\subsection{RKHS}

The concept of Reproducing Kernel Hilbert Spaces (RKHS) is widely used in statistics and machine learning (\cite{hofmann2008kernel}, \cite{gretton2012kernel}, \cite{sejdinovic2013equivalence}).

\begin{definition}[Reproducing Kernel Hilbert Space]
Let $\mathcal{X}$ be a non-empty set and let $\mathcal{H}$ be a Hilbert space of functions $f: \mathcal{X} \to \mathbb{R}$ with inner product $\langle \cdot, \cdot \rangle_\mathcal{H}$. A function $k: \mathcal{X} \times \mathcal{X} \to \mathbb{R}$ is called a reproducing kernel of $\mathcal{H}$ if it satisfies 
\begin{itemize}
    \item $k(\cdot, x) \in \mathcal{H}\quad \forall x \in \mathcal{X}$,
    \item (the reproducing property) $\langle f, k(\cdot, x) \rangle_\mathcal{H} = f(x) \quad \forall x \in \mathcal{X}, \forall f \in \mathcal{H}.$
\end{itemize}
If $\mathcal{H}$ has a reproducing kernel, then it is called reproducing kernel Hilbert space (RKHS).
\end{definition}

Based on the Cauchy-Schwarz inequality, the reproducing property implies that the map $F_x: f \in \mathcal{H} \to f(x)$ is a continuous linear form for all $x$ in $\mathcal{X}$. Therefore, the reproducing property introduces some degree of smoothness relative to the Hilbert space inner product of the functions considered (\cite{gretton2013introduction}).

\textbf{Notation:} Given a non-empty set $\mathcal{X}$, we refer to the considered kernel associated to $\mathcal{X}$ as $k_x$.  Given $x \in \mathcal{X}$, the element $k(\cdot, x) \in \mathcal{H}$ is also referred as $\phi_x(x)$, motivated by the notation used when understood as a feature representation of $x$. Given a data vector $\textbf{x} = [x_1, ..., x_n]^T$, feature matrices are defined by stacking feature maps along the columns $\Phi_{\bfx} := [\phi_x(x_1), ..., \phi_x(x_n)]$. The Gram matrix is denoted as $K_{\bfx \bfx} = \Phi_{\bfx}^T \Phi_{\bfx}$, and the vector of evaluations as $k_{x \bfx} = [k_x(x, x_1), ..., k_x(x, x_n)]$. Furthermore, $\Phi_{\bfx}(x) = k_{x \bfx}^T$. The notation is analogously used for the rest of variables used, such as $k_y$, $\textbf{y}$, $\Phi_{\bf{y}}$ or $K_{\bf{y} \bf{y}}$ for variable $y \in \mathcal{Y}$.

The Kernel Mean Embedding (KME), which allows for distribution representation based on RHKS, is the cornerstone of the algorithms that will be discussed in this work.  
 
\begin{definition}[Kernel Mean Embedding (KME)]
Let $\mathcal{P}$ be the set of all probability measures on a measurable space $(\mathcal{X}, \mathcal{B}_\mathcal{X})$, and $k: \mathcal{X} \times \mathcal{X}$ a reproducing kernel with associated RKHS $\mathcal{H}$ such that $sup_{x \in \mathcal{X}} k(x,x) < \infty$. The kernel mean embedding (KME) of $\mathbb{P} \in \mathcal{P}$ with respect to $k$ is defined as the following Bochner integral
$$\mu: \mathcal{P} \to \mathcal{H}, \quad \mathbb{P} \to \mu_\mathbb{P}:= \int k(\cdot, x) d\mathbb{P}(x).$$
\end{definition}
 
Kernel $k$ is said to be characteristic if $\mu$ in injective. In such case, $\mu_\mathbb{P}$ serves as a representation of $\mathbb{P}$. The frequently used Gaussian (RBF), Matérn and Laplace kernels are characteristic. The notion of characteristic kernel is an analogy to the characteristic function (\cite{fukumizu2007kernel}). The succeeding lemma immediately follows from the definition of kernel mean embedding. 
\begin{lemma}
The kernel mean embedding of $\mathbb{P}$ maps functions $f \in \mathcal{H}$ to their mean with respect to $\mathbb{P}$ through the inner product:
\begin{equation} \label{eq:kme_mean}
    \langle \mu_{\mathbb{P}}, f \rangle_\mathcal{H} = 
    \mathbb{E}_{X \sim \mathbb{P} }[f(X)].
\end{equation}
\end{lemma}
The conditional mean embedding $\mu_{Y | X = x}$ considers the kernel mean embedding of the conditional distribution $\mathbb{P}_{Y | X = x}$ for every $x \in \mathcal{X}$.

\subsection{Bayesian Conditional Mean Embedding}

A Bayesian learning framework on conditional embeddings was proposed in \cite{bayesimp}. In such framework, the conditional mean embedding $\mu_{Y | X = x}(y)$ is modeled by a Gaussian process $\mu_{gp}(x, y)$. A prior is defined over $\mu_{gp} \sim GP(0, k_x \otimes r_y)$, and the posterior models $\mu_{Y | X = x}(y)$.

Please note that the paths $\mu_{gp}(x, \cdot)$ should live in the RKHS associated to kernel $k$, as they model $\mu_{Y|X = x}(\cdot)$. If $r_y = k_y$, then the paths live outside the RKHS with probability 1. In order to ensure that the paths of such Gaussian process $\mu_{gp}(x, y)$ live almost surely within the RKHS, the prior covariate kernel may be defined as a nuclear dominant kernel over $k_y$ (\cite{sample_path_rkhs}). Following a similar structure to \cite{flaxman_bayesian_kernel_embedding} and \cite{bayesimp}, we choose $r_y$ to be the convolution of the original kernel with itself. Therefore, the GP prior over $\mu_{gp}$ is defined as follows:
\begin{equation}
\mu_{gp} \sim GP(0, k_x \otimes r_y),
\end{equation}
\begin{equation}
r_y(y_1, y_2) := \int k_y(y_1, u) k_y(u, y_2) \nu (du),
\end{equation}
where $\nu$ is a finite measure on $\mathcal{Y}$. The following function-valued regression is then set up:
\begin{equation}
    \phi_y(y_i) = \mu_{gp}(x_i, \cdot) + \lambda^{1/2} \epsilon_i,
\end{equation}
where $\epsilon_i \stackrel{iid}{\sim} GP(0, r)$ are noise functions. The noise hyperparameter of the GP links this framework to the spectral kernel mean shrinkage estimator (\cite{muandet_steins}). The posterior mean and covariance for $\mu_{gp}$, the marginal likelihood, and specific nuclear dominant kernels may be obtained in closed form and we refer to \cite{bayesimp} for the respective derivations.

\subsection{CGP-UCB}

In CGP-UCB, the reward sequence is assumed to be a sample from a known GP distribution. The exploration-exploitation trade-off translates consequently to a Bayesian optimization problem. At round $t$, the CGP-UCB picks action $a_t$ such that 
\begin{equation} \label{eq:ucb}
    a_t = argmax_{a \in A} \; \mu_{t-1}(s_t, a) + \beta_t^{\frac{1}{2}} \sigma_{t-1}(s_t, a),
\end{equation}
where $\beta_t$ are appropriate constants, and $\mu_{t-1}(\cdot)$ and $\sigma_{t-1}(\cdot)$ are the posterior mean and standard deviation of the GP over the joint set $S \times A$ conditioned on the observations $(s_1, a_1, r_1), ..., (s_{t-1}, a_{t-1}, r_{t-1})$. The addition 
\begin{equation} \label{eq:acq_function}
    \mu_{t-1}(s_t, a) + \beta_t^{\frac{1}{2}} \sigma_{t-1}(s_t, a)
\end{equation} is known as the acquisition function. CGP-UCB attains sublinear contextual regret in many real life applications. In other words, it is able to compete with the optimal mapping from contexts to actions.

\subsection{CME-UCB}

The Conditional Mean Embeddings Upper Confidence Bound (CME-UCB) algorithm was proposed in \cite{chowdhury2020active} for addressing the problem of sequentially maximising the conditional expectation of a function in an RKHS, using conditional mean embeddings. \cite{chowdhury2020active} defined the UCB acquisition function 
\begin{equation}
    \mu_{t-1}(a_t) + \beta_t^{\frac{1}{2}} \sigma_{t-1}(a_t),
\end{equation} 
where $\beta_t$ are appropiate constants,
\begin{equation}
    \mu_{t-1}(a) = \Phi(a)^{\top} (K_{\bfa \bfa} + \lambda I)^{-1} K_{\bfr \bfr}  (K_{\bfr \bfr } + \lambda_f I)^{-1} y,
\end{equation}
the standard deviation is described (as an application of the Sherman-Morrison formula) in terms of the Mahalanobis norm of the control features $\phi(a_t)$:
\begin{equation}
    \sigma_{t-1}(a) = \lambda^{-1/2} \sqrt{k_a(a, a) - \Phi_\bfa(a)^{\top} (K_{\bfa \bfa} + \lambda I)^{-1} \Phi_\bfa(a)}.
\end{equation}

Although this algorithm is not meant to deal with contexts and unmatched data, it is easily generalized. Given  $D_1 = \{ (s_i, a_i, r_i) \}_{i = 1}^{t-1}$, $D_2 = \{ (\tilde{r}_j, y_j) \}_{j = 1}^M$, and context $s_t$, it suffices to consider the following mean and standard deviation:
\begin{equation}
    \mu_{t-1}(a) = \Phi(s_t, a)^{\top} (K + \lambda I)^{-1} K_{\bfr \tilde{\bfr}}  (K_{\tilde{\bfr} \tilde{\bfr} } + \lambda_f I)^{-1} y,
\end{equation}
\begin{align}
    \sigma_{t-1}(a) = \lambda^{-1/2} \sqrt{G_{s_t, a}},
\end{align}
where $G_{s_t, a} =  \Phi(s_t, a)^\top \Phi(s_t, a) - \Phi(s_t, a)^{\top} (K + \lambda I)^{-1} \Phi(s_t, a)$, $K = K_{\bfs \bfs} \odot K_{\bfa \bfa}$ and $\Phi(s, a) = \Phi_{\bf{s}}^T k_s(s, \cdot) \odot \Phi_{\bf{a}}^T k_a(a, \cdot)$. Then, $a_t$ is chosen to maximize the acquisition function defined in Equation \eqref{eq:acq_function} with the corresponding $\mu_{t-1}$ and $\sigma_{t-1}$. Please note that the standard deviation $\sigma_t$ only depends on contexts and actions, without taking into consideration any information about $\{ \bfr, \tilde{\bfr}\}$.

\section{PROBLEM STATEMENT}

The problem addressed in this work is formally determined by a tuple $\langle \mathcal{S}, \mathcal{A}, \mathcal{R}, \mathcal{P}_R, F_Y \rangle$, where:
\begin{itemize}
    \item $\mathcal{S}$ is the space of contexts.
    \item $\mathcal{A}$ is the space of actions.
    \item $\mathcal{R}$ is the state of intermediate rewards.
    \item $\mathcal{P}_R: \mathcal{A} \times \mathcal{S} \times \mathcal{R} \to [0, 1]$ is the intermediate reward distribution.
    \item $F_Y: \mathcal{R} \times \mathbb{R} \to [0, 1]$ is the ultimate reward distribution.
\end{itemize}

At time step $t \in \mathbb{R}$, a context $s_t$ is given and the agent decides to take an action $a_t \in A$. Two subsequent rewards $r_{s_t, a_t}$ and $y_{r_t}$ drawn from $\mathcal{P}_R(\cdot | s_t, a_t)$ and $F_Y(\cdot | r_t)$ follow, which we name \textit{intermediate reward} and \textit{ultimate reward} respectively. Please note that we have not imposed any restriction to the space of intermediate rewards $\mathcal{R}$, however the ultimate reward is taken as real-valued. The RCMAB aims at maximizing the cumulative reward
\begin{equation}
    J_T = \mathbb{E}\left[\sum_{t = 1}^T y_{r_t}\right] = 
    \mathbb{E}\left[\sum_{t = 1}^T y_{r_{s_t, a_t}}\right],
\end{equation}
where $(s_1, ..., s_T)$ is sequentially given and $(a_1, ..., a_T)$ are the actions taken.

This problem is referred in the literature as the problem of sequentially maximizing the conditional expectation of a function. In \cite{chowdhury2020active}, the distribution $F_Y$ is expressed as a sum of the conditional mean and noise such that $y_t = f(r_t) + \zeta_t$. In contrast to \cite{chowdhury2020active}, we consider the intermediate reward $r_t | a_t, s_t \sim \mathcal{P}_R(a_t, s_t)$ to depend on both a context and action, rather than only an action i.e. we consider the contextualized version.

We highlight the importance of accounting for the whole distribution of the intermediate reward $\mathcal{P}_R$ in such framework. Even if we assume a simple deterministic relationship between intermediate and ultimate reward, the expectation of the ultimate reward cannot be maximized in terms of the expectation of the intermediate reward.

One may consider the reduced sample  $(s_t, a_t, y_t)_{t=1}^T$ by dropping the mediating variables $r_i$. However, information is disregarded, which could translate to a drop in the performance of algorithms (\cite{chowdhury2020active}). Furthermore, dropping $r_i$ does not allow to consider a second, unmatched data set $(r_j, y_j)_{j = 1}^M$. Hence such approach limits the power of the sequential decision making algorithms by disregarding information in case a second, unmatched data set is available.

We call attention to the different scopes of this work and \cite{bayesimp}. While in \cite{bayesimp} estimated interventional distributions are used to warm-start Bayesian optimization, the problem setting is itself static. By considering a multi-armed bandit framework rather than a causal inference problem, we make use of the uncertainty quantification proposed in \cite{bayesimp} for addressing a sequential exploration-exploitation trade-off. Furthermore, an extra term to be conditioned on (the context) is needed in this case, however no intervention needs to be accounted for.

\section{ALGORITHM DESIGN}

We propose a novel algorithm for tackling the problem of sequentially maximising the expectation of a function. It combines the ideas of the BayesIMP algorithm proposed in (\cite{bayesimp}) and the CGP-UBC. The mean and variance for each action are modeled following the ideas of BayesIMP. Then, the optimization of the CGP-UCB acquisition function follows, as the uncertainty estimates considered by the GPs can be used for such exploitation-exploration trade-off.

Formally, let $D_1 = (s_t, a_t, r_t)_{t = 1}^T$ be the sequential sample obtained from the interaction of the agent with the system, and let $D_2 = (r_j, y_j)_{j = 1}^{T+M}$ be the sample containing observations of the intermediate and ultimate rewards. As highlighted in the previous section, the whole distribution $\mathcal{P}_R(\cdot | s, a)$ ought to be modeled. As only the expectation of the final reward is of interest in the framework, there is no need to model the whole distribution $F_Y(\cdot | r_t)$. It suffices to estimate its expectation $\mathbb{E}[y_{r_t}]$, which we denote $f(r_t) = \mathbb{E}[y_{r_t}]$.

Based on these considerations and motivated by the BayesIMP algorithm (\cite{bayesimp}), we propose the following approach for the problem:

\begin{itemize}
    \item \textbf{Contextual Bayesian Mean Process UCB (CBMP-UCB)}: Expectation $f(r_t)$ is trained as a GP, and the conditional mean embedding of $\mathcal{P}_R(\cdot | s, a)$ is taken as a GP by considering the Bayesian conditional mean embedding $\mu_{R | s, a}$. A nuclear dominant kernel is needed in both $f$ and $\mu_{R | S = s, A = a}$ so that the inner product $\langle f, \mu_{R | s, a} \rangle$ can be considered almost surely. Although the process $\langle f, \mu_{R | s, a} \rangle$ is not a GP, CBMP-UCB picks action $a_t$ following Equation \eqref{eq:ucb} by taking $\mu_{t-1}(\cdot)$ and $\sigma_{t-1}(\cdot)$ as the mean and stardard deviation of $\langle f, \mu_{R | s_t, a} \rangle$ (i.e. moment matching is used to construct a GP out of $\langle f, \mu_{R | s, a} \rangle$ for posterior inference). 
\end{itemize}

The means and standard deviations, $\mu_{t-1}(\cdot)$ and $\sigma_{t-1}(\cdot)$, can be obtained in closed form. They have been developed in \cite{bayesimp} for a causal data fusion problem. As such problem requires adjusting for confounding variables, the closed forms expressions exhibited in \cite{bayesimp} are slightly more complex, although an extra term to be conditioned on (the context) is needed in this case. The following theorem could be interpreted as a simplified version of the BayesIMP theorem from \cite{bayesimp}. It is stated here for completeness with notation adapted for our setting and we refer to \cite{bayesimp} for its proof. 

\begin{proposition} 
Let  $D_1 = \{ (s_i, a_i, r_i) \}_{i = 1}^{t-1}$, $D_2 = \{ (\tilde{r}_j, y_j) \}_{j = 1}^M$ be two unmatched datasets and $k_s, k_a, k_r$ the kernels associated to variables $s, a, k$ respectively, and $r_r$ a nuclear dominant kernel of $k_r$.
We denote $K = K_{\bfs \bfs} \odot K_{\bfa \bfa}$, $\Phi(s, a) = \Phi_{\bf{s}}^T k_s(s, \cdot) \odot \Phi_{\bf{a}}^T k_a(a, \cdot)$, and $\hat{\bfr} = (\bfr, \tilde{\bfr})$.
If $f$ and $\mu_{R | s_t, a}$ are modeled as GP, then $g = \langle f, \mu_{R | s_t, a} \rangle$ has the following mean $m_3$ and variance $\kappa_3$:
\begin{align}
    m_3(a) &= E_{a} K_{\bfr \bf\hat{r}} K_{\bf\hat{r} \bf\hat{r}}^{-1} R_{\bf\hat{r} \bf\tilde{r}} (R_{\bf\tilde{r} \bf\tilde{r}} + \lambda_f I)^{-1} y, \\
    \kappa(a, a') &= E_a \Theta_1^{\top} \hat{R}_{\bf\hat{r} \bf\hat{r}} \Theta_1 E^{\top}_{a'} + \\
    &+ \Theta_2^{(a)} F_{a a'} - \Theta_2^{(b)} G_{a a'}  + \\ &+ \Theta_3^{(a)} F_{a a'} - \Theta_3^{(b)} G_{a a'},
\end{align}
where $E_{a} = \Phi(s_t, a)^{\top}(K + \lambda I)^{-1}$, 
$F_{a a'} = \Phi(s_t, a)^\top \Phi(s_t, a)$, 
$G_{a a'} = \Phi(s_t, a)^{\top}(K + \lambda I)^{-1} \Phi(s_t, a')^{\top}$, 
$\Theta_1 = K_{\bf\hat{r} \bf\hat{r}}^{-1} R_{\bf\hat{r} \bfr} R_{\bfr \bfr}^{-1} K_{\bfr \bfr}$, 
$ \Theta_2^{(a)} = \Theta_4^{\top} R_{\bf\hat{r} \bf\hat{r}} \Theta_4$,
$\Theta_2^{(b)} = \Theta_4^{\top} R_{\bf\hat{r} \bfr} R_{\bfr \bfr}^{-1} R_{\bfr \bf\hat{r}} \Theta_4$,
$\Theta_3^{(a)} = tr(K_{\bf\hat{r} \bf\hat{r}}^{-1} R_{\bf\hat{r} \bf\hat{r}} K_{\bf\hat{r} \bf\hat{r}}^{-1} \bar{R}_{\bf\hat{r} \bf\hat{r}})$, 
$\Theta_3^{(b)} = tr(R_{\bf\hat{r} \bfr} R_{\bfr \bfr}^{-1} R_{\bfr \bf\hat{r}} K_{\bf\hat{r} \bf\hat{r}}^{-1} \bar{R}_{\bf\hat{r} \bf\hat{r}} K_{\bf\hat{r} \bf\hat{r}}^{-1})$, 
$ \Theta_4 = K_{\bf\hat{r} \bf\hat{r}}^{-1} R_{\bf\hat{r} \bf\tilde{r}} (K_{\bf\tilde{r} \bf\tilde{r}} + \lambda_f)^{-1} y$. $\bar{R}_{\bf\hat{r} \bf\hat{r}}$ is the posterior covariance of f evaluated at $\hat{\bfr}$.

\end{proposition}

Interestingly and as explained in \cite{bayesimp}, the covariance of this closed form can be interpreted as the sum of covariances associated to $D_1$ and $D_2$, plus an interaction term between the two.

Please note that CMBP-UCB takes into account uncertainties derived from both $D_1$ and $D_2$ by considering a GP for modeling both expectation $f$ and conditional mean embedding $\mu_{R | s, a}$. In contrast, we recall that the CME-UCB algorithm (\cite{chowdhury2020active}) does not take into consideration any information of $\{ \bfr, \hat{\bfr} \}$ in the calculation of the standard deviation.

CBMP-UCB brings into play several hyperparameters: the lengthscales associated to the kernels, the noises of the GP, and regularization terms. Hyperparameter optimization can be conducted as often as possible, taking into consideration the computational resources available.

\section{EXPERIMENTS}
\begin{figure*}[t] 
  \includegraphics[width=\textwidth,height=14cm]{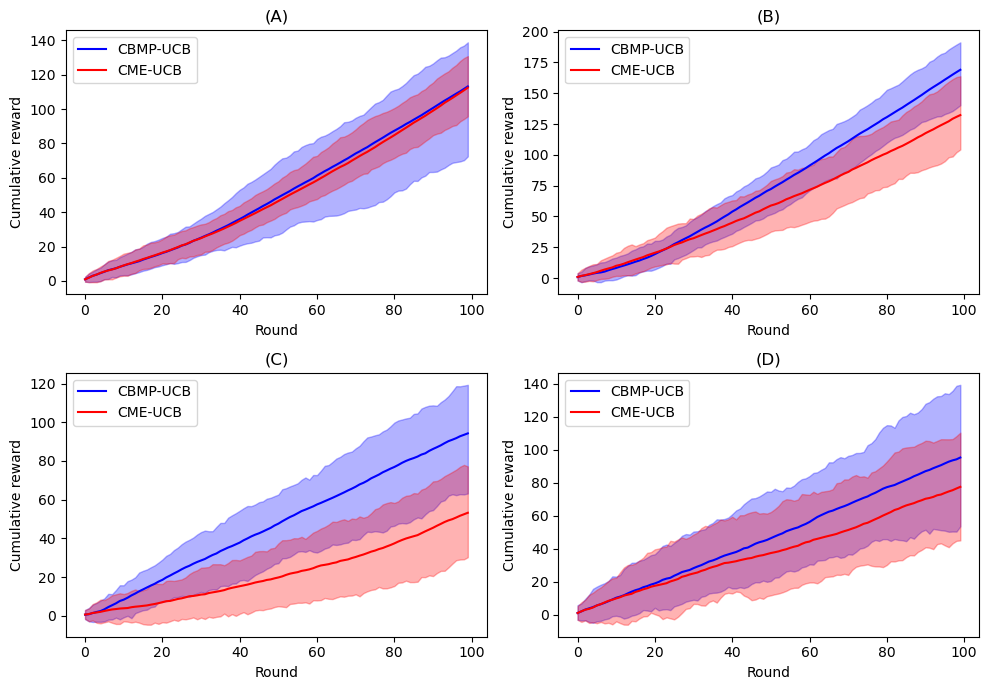}
  \caption{Mean and 0.05 quantiles of the cumulative rewards over 100 trials for settings A, B, C, and D with m = 0.}\label{fig:m0}
\end{figure*}

\begin{figure*}[t] 
  \includegraphics[width=\textwidth,height=14cm]{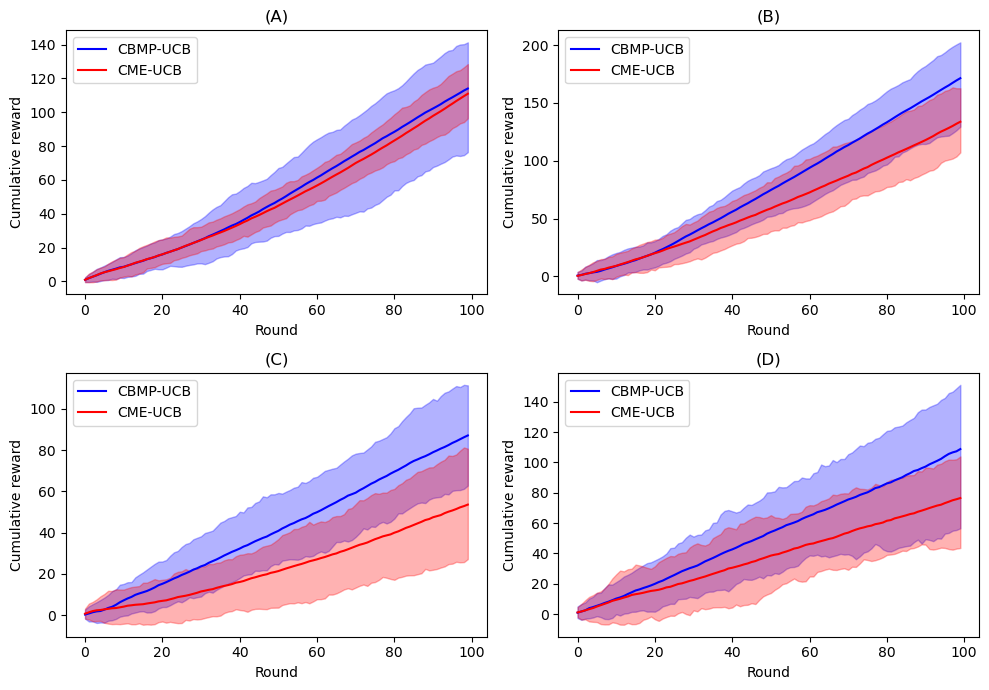}
  \caption{Mean and 0.05 quantiles of the cumulative rewards over 100 trials for settings A, B, C, and D with m = 50.}\label{fig:m30}
\end{figure*}

The main goal of the experiments conducted is to compare the performance of CBMP-UCB and CME-UCB. For this purpose, four experimental settings (A, B, C, D) are considered. Settings were selected to explore a range of different dependencies. 

Setting A considers a toy example with a one dimensional intermediate reward. Settings B and C contemplate a more complex, two-dimensional intermediate reward. In setting D, the dimension of the intermediate reward is raised to five. Furthermore, the conditional expectations differ in all four scenarios. 
\begin{itemize}
\item [i.] Setting A:
\begin{itemize}
    \item $\tilde{r}_j \sim Uniform(-2, 2)$.
    \item $y_j | r_j \sim 1.5 \sin{r_j} + 1 + 0.05 \mathcal{N}(0, 1)$.
    \item $s_i \sim Uniform (-3, 3.25)$.
    \item $r | s, a \sim \sin(\pi s) + \cos(\pi a) + 0.25\mathcal{N}(0, 1)$.
    \item $\mathcal{A} = [-3, 3.25]$, discretized in 61 points.
    \item $\mathcal{S} = [-3, 3.25]$, discretized in 61 points.
\end{itemize}

\item [ii.] Setting B:
\begin{itemize}
    \item $\tilde{r}_j \sim Uniform(-2, 2)^2$.
    \item $y_j | r_j = (r_j^0, r_j^1) \sim 1.5 (\tanh{(5 \pi r_j^0)} + \cos{(\pi r_j^0)}) + 1 + \mathcal{N}(0, 1) $.
    \item $s_i \sim Uniform (-3, 3.25)$.
    \item $r | s, a \sim [\sin{(s + a)} + 0.25\mathcal{N}(0, 1), \sin{(s - a)} + 0.25\mathcal{N}(0, 1)]$.
    \item $\mathcal{A} = [-3, 3.25]$, discretized in 61 points.
    \item $\mathcal{S} = [-3, 3.25]$, discretized in 61 points.
\end{itemize}

\item [iii.] Setting C:
\begin{itemize}
    \item $\tilde{r}_j \sim Uniform(-2, 2)^2$.
    \item $y_j | r_j = (r_j^0, r_j^1) \sim 1.5 (\sin{(\pi r_j^0)} + \sin{(\pi r_j^1)}) + 1 + \mathcal{N}(0, 1) $.
    \item $s_i \sim Uniform (-3, 3.25)$.
    \item $r | s, a \sim [\sin{(s)} + 0.25\mathcal{N}(0, 1), \cos{(a)} + 0.25\mathcal{N}(0, 1)]$.
    \item $\mathcal{A} = [-3, 3.25]$, discretized in 61 points.
    \item $\mathcal{S} = [-3, 3.25]$, discretized in 61 points.
\end{itemize}

\item [iv.] Setting D:
\begin{itemize}
    \item $\tilde{r}_j \sim Uniform(-2, 2)^5$.
    \item $y_j | r_j = (r_j^0, r_j^1, r_j^2, r_j^3, r_j^4) \sim 1.5 (\sin{(\pi r_j^0)} + \sin{(\pi r_j^1)} + \sin{(\pi r_j^2)} + \sin{(\pi r_j^3)} + \sin{(\pi r_j^4)}) + 1 + \mathcal{N}(0, 1) $.
    \item $s_i \sim Uniform (-3, 3.25)$.
    \item $r | s, a \sim [\sin{(s)}, \cos{(a)}, \sin{(a)}, \cos{(s)}, \sin{(s)} + \cos{(s)}] + 0.25\mathcal{N}(0, I_5)$.
    \item $\mathcal{A} = [-3, 3.25]$, discretized in 61 points.
    \item $\mathcal{S} = [-3, 3.25]$, discretized in 61 points.
\end{itemize}

\end{itemize}

Kernels $k_s$, $k_a$ and $k_r$ were taken as RBF kernels for the CBMP-UCB algorithm. For the implementation of the CME-UCB algorithm, the choice of the kernels considered was motivated by \cite{chowdhury2020active}: $k_s$ and $k_a$ were taken as Mátern kernels, and $k_r$ as an RBF kernel. 

For each setting, we considered both a \textit{matched} and an \textit{unmatched} scenario, where $m = 0$ and $m = 50$ respectively. We run 100 trials with different random seeds for the CME-UCB and CBMP-UCB algorithms for 100 rounds.

Figure \ref{fig:m0} exhibits the results of the \textit{matched} case, in terms of the means and the $0.05$ quantiles. We observe that the mean of the cumulative reward of both CBMP-UCB and CME-UCB is very similar in the one-dimensional intermediate reward experiment (setting A), although the variance of CME-UCB seems smaller. In contrast, CBMP-UCB outperforms CME-UCB in the other three settings considered, where the dimension of the intermediate reward is higher. In setting B and setting C, both algorithms present a similar variance, although CBMP-UCB clearly outperforms CME-UCB. The variance of the cumulative reward of both algorithms goes up in setting D, which is explained by the higher complexity of the intermediate reward. Although the CBMP-UCB algorithm shows slightly more variance, it still outperforms CME-UCB. 

Figure \ref{fig:m30} displays the results of the \textit{unmatched} case, where $m = 50$. The results look very similar to those in Figure \ref{fig:m0}. The main difference between the performance of the algorithms with $m = 0$ and $m = 50$ is shown in setting D, in which additional information on the dependence between the five dimensional intermediate reward and the ultimate reward is crucial, given the higher complexity of the space of intermediate rewards. CBMP-UCB seems to be benefiting more from the initial information on such dependence when $m = 50$, given that the difference in performance is greater in this scenario. It is consistent with the fact that CBMP-UCB uses information of $D_2$ for the variance estimates, in contrast to CME-UCB.

Please note that, in general, CBMP-UCB outperforms CME-UCB in the settings considered where the intermediate reward is complex. Given the fact that only CBMP-UCB factors in the uncertainty arising from $D_2$, the illustrations of the two methods back the idea that considering uncertainties stemming from the two data sets is desirable, at least when the behaviour of the intermediate reward is complex enough. Considering uncertainty derived from $D_2$ does not prove to be useful in Experiment A, where the potential exploration of the one dimensional intermediate reward is limited by nature. Experiments B, C and D show a significant improvement in the performance if taking into account the uncertainty stemming from $D_2$. Such reasoning is consistent with the theoretical background, as not considering the uncertainty inherited by the lack of information on $f$ may imply a lack of exploration of regions of the space of intermediate rewards, which may be especially detrimental when the distribution of intermediate rewards is complex.

\section{CONCLUSION}

We have designed a novel algorithm CBMP-UCB for tackling the problem of sequentially maximising the expectation of a function based on the ideas of BayesIMP (\cite{bayesimp}) and CGP-UCB (\cite{krause2011contextual}), allowing for contexts and unmatched data sets. In the experiments considered, the CBMP-UCB has surpassed the overall performance of the baseline CME-UCB algorithm (\cite{chowdhury2020active}), especially when the dimensionality of the intermediate reward is raised. Such conclusion is consistent with the theoretical background, as disregarding the uncertainty stemming from the lack of information on $f$ may imply an insufficient exploration of the space of intermediate rewards.

There are several lines of research that could naturally follow this work. First of all, we have mentioned that hyperparameter optimization may be conducted as often as desired, given the computational limitations. We highlight that conducting hyperparameter optimization severely slows the algorithm, especially when having abundant data. Trying to determine a strategy on the rate of hyperparameter optimization could be addressed in future work. 

Furthermore, only the expectation of the ultimate reward has been considered in the proposed framework. However, one may instead be interested in other properties of the distribution of the ultimate reward, for which modeling the full distribution via its embedding as a GP could be explored. Considering several embeddings would potentially enable generalizing the methods so that they allow for a nested sequence of intermediate rewards, which could be explored in the future.

Lastly, we highlight that all the algorithms presented require the inversion of data matrices. Similarly to most kernel methods, the algorithms suffer from scalability issues inherited by this inversion. Such problem inevitably manifests when the agent has interacted for many rounds with its environment in the RCMAB framework. Large-scale approximations to kernel matrices are now well established (\cite{li2021towards}) and can be explore in our context. A deep learning approach (\cite{collier2018deep}) is another potential direction following this work.

\subsubsection*{Acknowledgements}
Diego Martinez-Taboada gratefully acknowledges the support provided by the Barrie Foundation.

\bibliographystyle{unsrtnat}
\bibliography{references}  %%% Uncomment this line and comment out the ``thebibliography'' section below to use the external .bib file (using bibtex) .





% \end{thebibliography}

\end{document}